\PassOptionsToPackage{prologue,dvipsnames,table}{xcolor}
\documentclass[letterpaper, 10 pt, conference, table]{ieeeconf}  

\IEEEoverridecommandlockouts                              

\pdfoutput=1

\usepackage[dvipsnames]{xcolor}
\usepackage[table]{xcolor}
\usepackage[T1]{fontenc}
\usepackage{cite}
\usepackage{graphicx}
\usepackage{amsmath, amsfonts}
\usepackage{tabularx, booktabs, utfsym, color, colortbl, arydshln, multirow, epsfig}
\usepackage{makecell, arydshln}

\definecolor{c1}{RGB}{189,230,205}
\definecolor{c2}{RGB}{228,238,188}
\definecolor{c3}{RGB}{255,248,197}
\definecolor{c4}{RGB}{238,238,238}

\title{\LARGE \bf
GLC-SLAM: Gaussian Splatting SLAM with Efficient Loop Closure
}

\author{Ziheng Xu$^{1}$, Qingfeng Li$^{1}$, Chen Chen$^{2}$, Xuefeng Liu$^{1}$ and Jianwei Niu$^{1*}$
\thanks{$^{1}$Ziheng Xu, Qingfeng Li, Xuefeng Liu and Jianwei Niu are with Beihang University, Beijing, China.
        }%
\thanks{$^{2}$Chen Chen is with the Hangzhou Innovation Institute of Beihang University, Hangzhou, China.
        }%
\thanks{*Corresponding author.}
}

\begin{document}

\maketitle
\thispagestyle{empty}
\pagestyle{empty}

\begin{abstract}

3D Gaussian Splatting (3DGS) has gained significant attention for its application in dense Simultaneous Localization and Mapping (SLAM), enabling real-time rendering and high-fidelity mapping. However, existing 3DGS-based SLAM methods often suffer from accumulated tracking errors and map drift, particularly in large-scale environments. To address these issues, we introduce GLC-SLAM, a Gaussian Splatting SLAM system that integrates global optimization of camera poses and scene models. Our approach employs frame-to-model tracking and triggers hierarchical loop closure using a global-to-local strategy to minimize drift accumulation. By dividing the scene into 3D Gaussian submaps, we facilitate efficient map updates following loop corrections in large scenes. Additionally, our uncertainty-minimized keyframe selection strategy prioritizes keyframes observing more valuable 3D Gaussians to enhance submap optimization. Experimental results on various datasets demonstrate that GLC-SLAM achieves superior or competitive tracking and mapping performance compared to state-of-the-art dense RGB-D SLAM systems.

\end{abstract}

\section{INTRODUCTION}

Visual SLAM plays a crucial role in various applications such as virtual/augmented reality (VR/AR), robot navigation, and autonomous driving. 
Over the past decade, visual SLAM methods with various scene representation have been developed, ranging from traditional approaches using point clouds \cite{18,33}, surfels \cite{29,30} and voxels \cite{31,32} to neural implicit methods \cite{4,10,7,25,34,38} leveraging neural radiance fields (NeRF) \cite{19}. Traditional SLAM methods provide accurate tracking and real-time mapping but struggle to generate high-quality, texture-rich maps or synthesize novel views. In contrast, NeRF-based SLAM methods offer coherent mapping and accurate surface reconstruction but are limited by the high computational cost of volume rendering, hindering real-time performance.

Recently, 3DGS \cite{23} has emerged as a promising alternative to NeRF, offering comparable high-quality rendering with significantly faster rendering and training speeds. Consequently, SLAM methods \cite{8,9,16,17,36,37} based on Gaussian Splatting representation demonstrate advancements in terms of photo-realistic rendering, high-fidelity reconstruction and real-time performance. It is worth noting that 3D Gaussian maps can be explicitly edited and deformed, making them particularly suitable for map correction.

However, existing 3DGS-based SLAM methods face the challenge of error accumulation and map distortion due to the absence of loop closure for global adjustment of camera poses and the constructed map. While Photo-SLAM \cite{17} incorporates loop closure based on ORB-SLAM \cite{18}, its dependence on a feature-based tracker constrains the effectiveness of loop closure, as the tracker is unable to exploit the map refinements. NeRF-based SLAM methods \cite{14,27,28} integrate online loop closure to achieve accurate and robust tracking, yet require storing historical frames and costly retraining the entire implicit map to update loop correction. The lack of a robust, efficient loop closure in 3DGS-based SLAM remains a key limitation to achieving global consistency in large-scale environments.

\begin{figure}[t]
    \begin{minipage}[c]{0.48\linewidth}		
        \centering\small\text{GO-SLAM \cite{14}}
    \end{minipage}
    \begin{minipage}[c]{0.48\linewidth}		
        \centering\small\text{Gaussian-SLAM \cite{9}}
    \end{minipage}
    
    \begin{minipage}[c]{0.48\linewidth}		
        \epsfig{figure=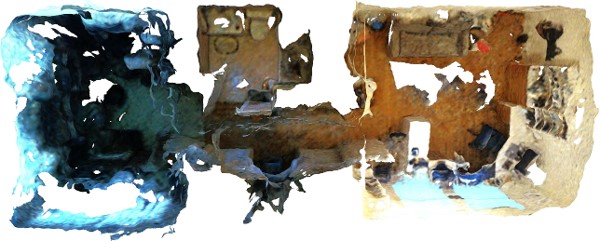,width=4cm}
    \end{minipage}
    \vspace{2pt}
    \begin{minipage}[c]{0.48\linewidth}		
        \epsfig{figure=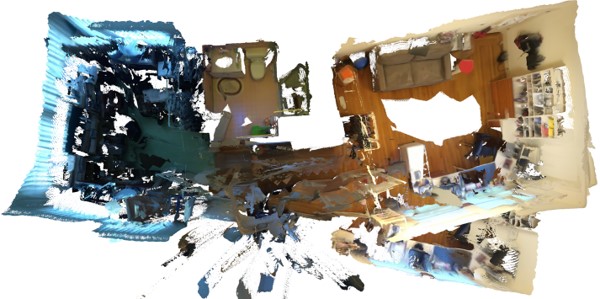,width=4cm}
    \end{minipage}
    \vspace{2pt}
    
    \begin{minipage}[c]{0.48\linewidth}		
        \centering\small\text{GLC-SLAM (Ours)}
    \end{minipage}
    \begin{minipage}[c]{0.48\linewidth}		
        \centering\small\text{GT}
    \end{minipage}
    
    \begin{minipage}[c]{0.48\linewidth}		
        \epsfig{figure=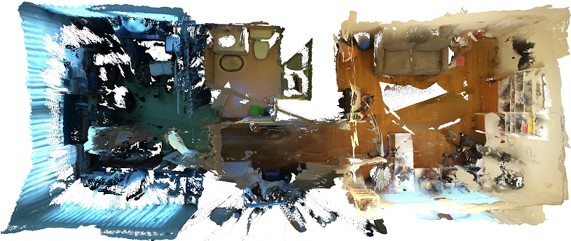,width=4cm}
    \end{minipage}
    \begin{minipage}[c]{0.48\linewidth}		
        \epsfig{figure=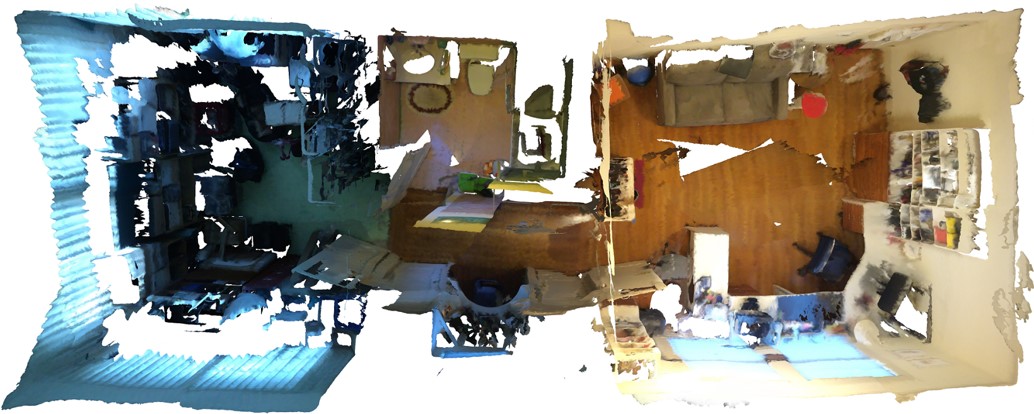,width=4cm}
    \end{minipage}
    \caption{\textbf{Reconstruction results on ScanNet \cite{3} \texttt{0054}}. Our method effectively mitigates the severe map drift inherent in Gaussian-SLAM \cite{9}, while also providing superior scene geometry and detail compared to GO-SLAM \cite{14}.}
    \label{fig:Figure1}
\end{figure} 

To address these challenges, we propose GLC-SLAM, a Gaussian Splatting SLAM system with efficient Loop Closure, designed to mitigate accumulated tracking errors and reduce map drift in large indoor scenes. Our approach incrementally builds 3D Gaussian submaps, with each submap anchored to a corresponding global keyframe. To maintain global consistency, we employ a hierarchical loop closing strategy that enhances global loop closure by drift-free submaps refined via local optimization. Upon loop detection, nodes and edges are added to the pose graph, followed by pose graph optimization. The optimization results are then updated to relevant submaps through direct map adjustment. Furthermore, we explicitly model Gaussian uncertainty and introduce an uncertainty-minimized keyframe selection method for robust active submap optimization. As shown in Fig. \ref{fig:Figure1}, GLC-SLAM successfully address map drift and yields improved scene geometry and detail, achieving high-fidelity and global consistent mapping. We conduct experiments on various datasets that demonstrate our method achieves robust tracking and accurate mapping performance compared to existing dense RGB-D SLAM methods.

Our main contributions are summarized as follows:
\begin{itemize}
    \item A Gaussian Splatting SLAM system that achieves robust frame-to-model tracking and global consistent mapping of 3D Gaussian submaps in large-scale environments.
    \item The efficient loop closure module, including global-to-local loop detection, pose graph optimization and direct map updates, reducing accumulated errors and map drift.
    \item The uncertainty-minimized keyframe selection strategy, which selects informative keyframes observing more stable 3D Gaussians during submap optimization to enhance mapping accuracy and robustness.
\end{itemize}

\section{RELATED WORK}
\subsection{Visual SLAM}
Early methods like ORB-SLAM \cite{33} utilize feature-based approaches to estimate camera trajectories and construct 3D maps. While traditional SLAM systems, which typically employ explicit representations like voxels and point clouds, excel in tracking accuracy and efficiency, they are limited in providing high-fidelity maps and often lack generalization capabilities.

In recent years, NeRF \cite{19} have gained significant attention in SLAM algorithms, with notable examples like iMAP \cite{4}, NICE-SLAM \cite{10}, and ESLAM \cite{7} leveraging neural implicit representations for accurate and dense 3D surface reconstruction. However, these neural implicit methods are constrained by the high computational demands of volume rendering and face challenges in performing robust tracking in large-scale environments. To improve tracking robustness, some approaches incorporate loop closure and online global bundle adjustment (BA) to mitigate accumulated error. For example, MIPS-Fusion \cite{13} employs a multi-implicit-submap representation, achieving global optimization by refining and integrating these submaps, while GO-SLAM \cite{14} combines loop closure with online full BA across all keyframes to ensure global consistency in large-scale environments. However, these methods either require storing the entire history of input frames or involve time-consuming retraining for map updates after loop closure.

\subsection{3DGS-based SLAM}
SLAM methods based on 3D Gaussian representation have recently garnered broad interest due to their ability to combine the strength of explicit and implicit expressions. Compared to NeRF-based methods, 3DGS-based methods capture high-fidelity 3D scenes through a differentiable rasterization process, avoiding the per-pixel ray casting required by neural fields, thus achieving real-time rendering. Gaussian-SLAM \cite{9} organize scenes as 3D Gaussian submaps, allowing for efficient optimization and preventing catastrophic forgetting. SplaTAM \cite{8} employs simplified 3D Gaussian representation, enabling real-time efficient optimization and high-quality rendering. However, these methods lack online loop correction, leading to the accumulation of errors and map drift. Photo-SLAM \cite{17}, building on ORB-SLAM \cite{18}, integrates loop closure to reduce cumulative errors and enhance tracking robustness, yet its design decouples tracking from mapping, which diminishes the effectiveness of loop closure and increases communication overhead. Our method constructs 3D Gaussian submaps incrementally and employs frame-to-model tracking, achieving a coupled SLAM system while reducing unnecessary storage consumption. By applying hierarchical loop closure and rapidly updates the scene through map deformation, we ensure robust tracking and efficient mapping in large-scale environments.

\section{PRELIMINARY}
\begin{figure*}[ht]
	\centering
		\includegraphics[width=0.98\linewidth]{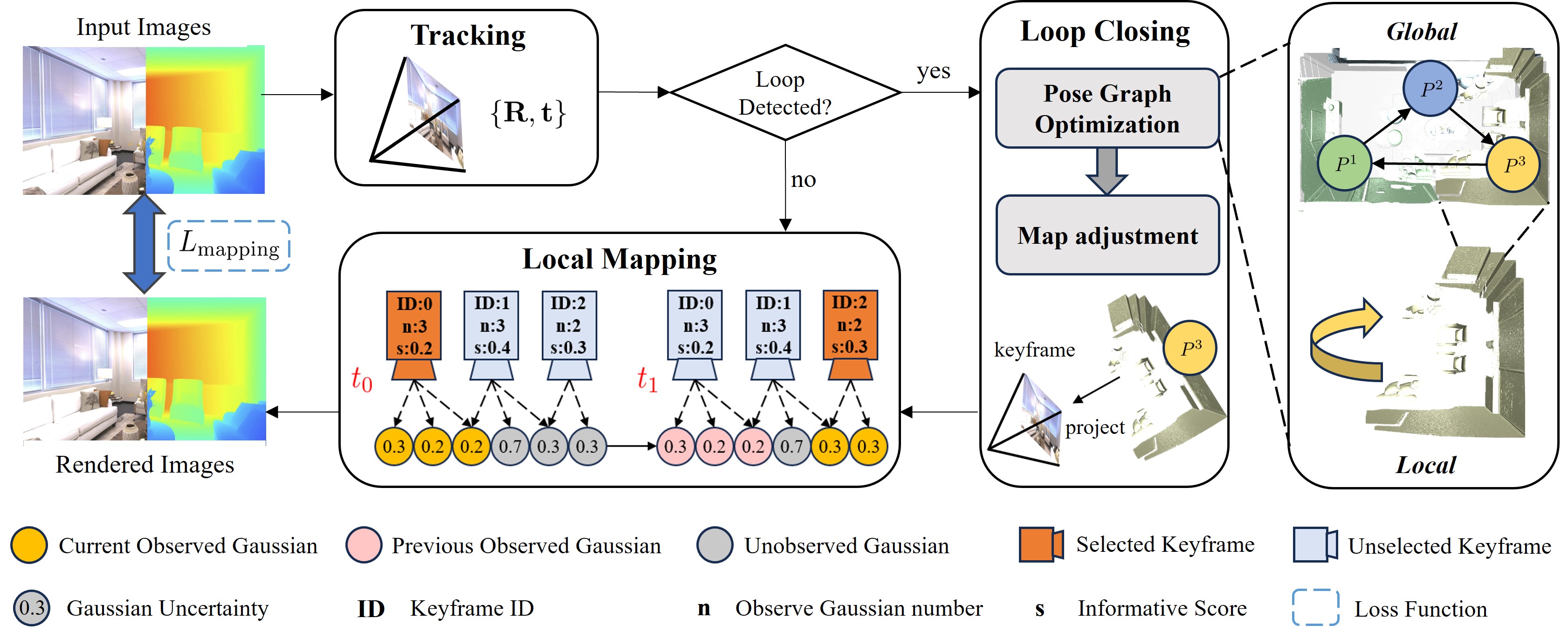}
	\caption{\textbf{System Overview.} Our system consists of three processes: tracking, mapping and loop closing. The tracking process estimates and refines camera poses $\{\text{R, t}\}$ by minimizing the tracking loss. The scene is managed as Gaussian submaps while the local mapping process select keyframes with an uncertainty-minimized strategy to optimize the active submap. If a loop is detected, the loop closing process triggers loop closure online, followed by efficient map adjustment to correct accumulated error and mitigate map drift.}
	\label{fig:Figure2}
\end{figure*}
\subsection{Scene Representation}
We represent the scene using 3D Gaussian submaps, where each submap $P^s$ consists of a collection of N 3D Gaussian distributions:
\begin{equation}
    P^s=\{G_i^s(\mu,\Sigma,o,C) \mid i=1,\dots,N\},
    \label{eq:1}
\end{equation}
each 3D Gaussian is parameterized by mean $\mu$, covariance $\Sigma$, opacity $o$, and RGB color $C$. The covariance matrix $\Sigma$ is decomposed into a rotation vector $r$ and a scale vector $s$. By using differentiable splatting to render color and depth maps, 3D Gaussians are optimized through an iterative process that involves calculating errors with input RGB-D images and updating the Gaussian parameters accordingly.

The color image $\hat{C}$ and depth map $\hat{D}$ can be rendered by alpha-blending proposed in \cite{23}:
\begin{equation}
\begin{split}
    \hat{C}=\sum_{i=1}^{n}& c_i\alpha_iT_i ,\hat{D}=\sum_{i=1}^{n} d_i\alpha_iT_i,\\
    &T_i=\prod_{j=1}^{i-1} (1-\alpha_j) ,
\end{split}
    \label{eq:2}
\end{equation}
where $c_i$ and $d_i$ are the color and depth value of a 3D Gaussian. $\alpha_i$ is computed by the pixel coordinate $u$, mean $\mu$ and covariance matrix $\Sigma_{2D}$ of the splatted 2D Gaussian in pixel space:
\begin{equation}
    \alpha_i=o_i \exp (-\frac{1}{2}(u-\mu)^T \Sigma_{2D}^{-1}(u-\mu)).
    \label{eq:3}
\end{equation}


\subsection{Uncertainty Modeling}
Uncertainty modeling introduces non-uniform weights to select more valuable pixels and 3D Gaussians during optimization, rather than treating them with equal importance. Following \cite{22}, we explicitly model the uncertainty of the rendered depth images and 3D Gaussians.

We render the depth uncertainty map $U$ as:
\begin{equation}
    U = \sum_{i=1}^{N} \alpha_iT_i(d_i-D)^2,
    \label{eq:4}
\end{equation}
where $D$ represents the ground truth depth values.

We define the dominated pixels of a 3D Gaussian same as \cite{22} and calculate the uncertainty $\nu_i$ of the $i$-th 3D Gaussian by the difference between its depth and depth observations from all its dominated pixels $P = \{p_1,p_2,\ldots,p_n\}$ within a keyframe window. 
\begin{equation}
    \nu_i=\frac{1}{n} \sum_{p_k \in P} \alpha_i^{k} T_i^{k}\left(D^k-d_i^k\right)^2 .
    \label{eq:5}
\end{equation}
$\alpha_i^{k} \text{ and } T_i^{k}$ represent the opacity and transmittance of the $i$-th 3D Gaussian on a pixel $p_k$. $D^k$ and $d_i^k$ represent the ground truth depth on a pixel $p_k$ and the depth value of the $i$-th 3D Gaussian respectively.

\section{SYSTEM}
The overview of our proposed GLC-SLAM system is shown in Fig. \ref{fig:Figure2}. In this section, we introduce our system from the following aspects: tracking (\ref{sec. 4.1}), local mapping (\ref{sec. 4.2}) and loop closing (\ref{sec. 4.3}).
\subsection{Tracking} \label{sec. 4.1}
We adopt a coupled system design by performing frame-to-model tracking based on the mapped scene. We first initialize the current camera pose $T_i$ with a constant speed assumption:
\begin{equation}
    T_i = T_{i-1}+(T_{i-1}-T_{i-2}) ,
    \label{eq:6}
\end{equation}
where camera pose $T_i=\{R_i, t_i\}$ can be decompose into a rotation matrix and a translation vector. $T_i$ is then optimized by minimize the tracking loss $L_\text{tracking}$ with respect to relative camera pose $T_{i-1,i}$ between frames $i-1$ and $i$. We apply an alpha mask $M_\text{alpha}$ and an inlier mask $M_\text{inlier}$ in the tracking loss to address gross errors caused by poorly reconstructed or previously unobserved areas as follows:
\begin{equation}
    L_{\text {tracking }}=\sum M_{\text {in}} \cdot M_{\text {alpha}} \cdot(\lambda_c|\hat{C}-C|_1+\left(1-\lambda_c\right)|\hat{D}-D|_1),
    \label{eq:7}
\end{equation}
where $\lambda_c$ is a weight that balances the color and depth losses, and $C$ and $D$ are the input color and depth map. 

\subsection{Local Mapping} \label{sec. 4.2}
We grow submaps of 3D Gaussians in a progressive manner and anchor each submap to a global keyframe. 
All Gaussians in the active submap are jointly optimized every time new Gaussians are added to the submap for a fixed number of iterations minimizing the loss Eq. (\ref{eq:13}) and only the selected keyframes are included in the optimization. 

\subsubsection{Map Building} We grow submaps incrementally with newly incoming keyframes and initialize new submaps when the camera motion exceeds a threshold, with the first keyframe serving as a global keyframe. At any time, only the active submap is processed. This approach bounds the compute cost and ensures that optimization remains fast while exploring larger scenes. 

Each new keyframe adds 3D Gaussians to the active submap, capturing newly observed regions of the scene. Specifically, a dense point cloud is computed from the RGB-D input following the pose estimation for the current keyframe. We apply a densification mask to fill holes of unobserved regions and avoid local minima in rendered images. Points are sampled uniformly from the regions with the accumulated alpha values lower than a threshold $\alpha_\text{thre}$ or large rendered color and depth error occurs. New 3D Gaussians are added to the submap using sampled points that have no neighbors within a search radius in the current submap. The new Gaussians are anisotropic and their scales are defined based on the nearest neighbor distance within the active submap.

\subsubsection{Uncertainty-minimized Keyframe Selection} 
For a new input frame, we insert the frame into the keyframe set if the frame overlap ratio $r_o$ between the current frame and the last inserted frame is lower than a threshold, where $r_o$ is defined as:
\begin{equation}
    r_o=\frac{G_i \cap G_{i-1}}{G_i \cup G_{i-1}} .
    \label{eq:8}
\end{equation}
Here, $G_i \text{ and } G_{i-1}$ are the 3D Gaussian sets observed by the current frame and last keyframe respectively.

Inspired by \cite{21}, we adopt an uncertainty-aware keyframe selection strategy in each map training iteration. This strategy, with the aid of Gaussian uncertainty, aims to select keyframes that observe more valuable 3D Gaussians which are likely to have a positive effect on the optimization. An informative score is defined for each keyframe as:
\begin{equation}
    s_{\mathrm{infor}}=\frac{1}{|G|}\sum_{g \in G}\nu_g ,
    \label{eq:9}
\end{equation}
where $|G|$ is the number of observed 3D Gaussians by the keyframe.

We begin by selecting $k$ keyframes that cover the Gaussians with the highest sum of scores. After labeling the covered Gaussians as observed we use the same selection strategy but only consider the remaining unobserved Gaussians when calculating $s_{\mathrm{infor}}$ in the next time step. If all Gaussians have been labeled as observed, the process is repeated by resetting the Gaussians to be labeled as unobserved.

\subsubsection{Loss Function}
We employ various loss functions to optimize Gaussian parameters. For depth supervision, we use the loss:
\begin{equation}
    L_{\text{depth}}=\frac{1}{U}\|D-\hat{D}\|_1,
    \label{eq:10}
\end{equation}
with $D$ and $\hat{D}$ being the ground-truth and reconstructed depth maps, respectively. The depth loss $L_\text{depth}$ is weighted by the uncertainty map $U$ to ensure that the pixels with high uncertainty are weighted less. For the color supervision we use a weighted combination of $L1$ and SSIM \cite{24} losses:
\begin{equation}
    L_{\text{color}}=(1-\lambda) \cdot|\hat{C}-C|_1+\lambda(1-\operatorname{SSIM}(\hat{C}, C)) \text {, }
    \label{eq:11}
\end{equation}
where $C$ is the original image, $\hat{C}$ is the rendered image, and $\lambda$ = 0.2. We also add an isotropic regularization term $L_\text{reg}$:
\begin{equation}
    L_{\text{reg}}=\frac{1}{|P|}\sum_{p \in P}|s_p-\bar{s}_p|_1
    \label{eq:12}
\end{equation}
where $P$ is a submap, $s_p$ is the scale of a 3D Gaussian, $\bar{s}_p $ is the mean submap scale, and $|P|$ is the number of 3D Gaussians in the submap. The final loss function for mapping is finally formulated as:
\begin{equation}
    L_\text{mapping}=\lambda_\text{color}\cdot L_\text{color}+\lambda_\text{depth}\cdot L_\text{depth}+\lambda_\text{reg}\cdot L_\text{reg}
    \label{eq:13}
\end{equation}
where $\lambda_\text{color},\lambda_\text{depth},\lambda_\text{reg}$ are weights for the corresponding losses.

\subsection{Loop Closing} \label{sec. 4.3}
We employ hierarchical loop closure to achieve global consistency within and between submaps. Global loop closure corrects large inter-submap cumulative errors while local loop closure aid global correction with refined global keyframe poses and accurate intra-submap geometry.

\subsubsection{Loop Detection}
For place recognition, we use the pre-trained NetVLAD \cite{1} model to extract a feature descriptor for each keyframe. The extracted features are stored in global and local keyframe databases. Cosine similarity between descriptors serves as the criterion for loop detection.

Global loop detection is triggered when a new submap is created. We select the best match from the global keyframe database if the visual similarity score is higher than a threshold $s_\text{global}$, which is dynamically computed as the minimum score between the global keyframe and the keyframes within active submap. Local loop detection operates during the local mapping process, accepting the most similar keyframe with the similarity score exceeds a predefined threshold $s_\text{local}$. To avoid false loops, especially in indoor scenes with repetitive objects like chairs or tables, we further apply a geometry check. We evaluate the frame overlap ratio between two loop candidate keyframes, and accept them if $r_o$ exceeds a threshold.

\subsubsection{Pose Graph Optimization}
We construct a pose graph model where the nodes represent keyframe poses, and the edges correspond to sequential relative poses. Loop edge constraints are computed from the relative poses between loop nodes and subsequently added to the pose graph.

We perform pose graph optimization across the entire pose graph to align the estimated trajectory more closely to the ground truth. Pose graph optimization effectively mitigates cumulative error and improves tracking accuracy. We use the Levenberg-Marquarelt algorithm to solve this nonlinear pose graph optimization problem described by Eq. (\ref{eq:14}), where $v$ is the set of nodes, $E_s$ is the set of sequential edges, $E_l$ is the set of loop edges and $\Lambda_i$ represents the uncertainty of corresponding edges.
\begin{equation}
    v^*=\arg \min _v \frac{1}{2} \sum_{e_i \in E_s, E_l} e_i^T \Lambda_i^{-1} e_i,
    \label{eq:14}
\end{equation}

\subsubsection{Map Adjustment}
To maintain map consistency after pose graph optimization, we rearrange the 3D Gaussian submaps using a keyframe-centric adjustment strategy. Each 3D Gaussian $g_i$ is associated to a keyframe, and submap adjustment is achieved by updating Gaussian means based on the optimized pose of the associated keyframe. Association is determined by which keyframe added the 3D Gaussian to the scene. The mean $\mu_i$ is projected into $T^\prime$ to find the pixel correspondence. Specifically, assume that a keyframe with camera pose $T=\{R, t\}$ is updated to $T^\prime=\{R^\prime, t^\prime\}$, we update the mean and rotation of all 3D Gaussians $g_i$ associated with the keyframe. We update $\mu_i$ and $r_i$ accordingly as:
\begin{equation}
    \boldsymbol{\mu}_i^{\prime}= T^{\prime} T^{-1} \boldsymbol{\mu}_i, r_i^{\prime}=R^{\prime} R^{-1} r_i.
    \label{eq:15}
\end{equation}

After map adjustment, we perform a set of refinement steps on the updated submap. We disable pruning and densification of the 3D Gaussians and simply perform a set of optimization iterations using the same loss function Eq. (\ref{eq:13}).

\begin{table}[ht]
\setlength\tabcolsep{3pt}
	\begin{center}
		\caption{\textbf{Tracking Performance on Replica \cite{1}}. The best results are highlighted as \colorbox{c1}{\textbf{first}} , \colorbox{c2}{second}, and \colorbox{c3}{third}. $^\ast$ indicates methods leveraging external tracker.}\label{tab:Table 1.}
		\resizebox{\linewidth}{!}{
			\begin{tabular}{lcccccccccc}
				\toprule
				\text{Method} & \texttt{rm0} & \texttt{rm1} & \texttt{rm2} & \texttt{off0} 
                & \texttt{off1} & \texttt{off2} & \texttt{off3} & \texttt{off4} & \text{Avg.} \\
				\midrule
                \multicolumn{10}{>{\columncolor{c4}}l}{\textit{NeRF-based}}\\
				NICE-SLAM \cite{10} & 0.97 & 1.31 & 1.07 & 0.88 & 1.00 & 1.06 & 1.10 & 1.13 & 1.06\\
			    Vox-Fusion \cite{11} & 1.37 & 4.70 &1.47 & 8.48 & 2.04 & 2.58 & 1.11 & 2.94 & 3.09\\
				ESLAM \cite{7} & 0.71 & 0.70 & 0.52 & 0.57 & 0.55 & 0.58 & 0.72 & 0.63 & 0.63\\
				Point-SLAM \cite{12} & 0.61 & 0.41 & 0.37 & 0.38 & 0.48 & 0.54 & 0.69 & 0.72 & 0.52\\
                MIPS-Fusion \cite{13} & 1.10 & 1.20 & 1.10 & 0.70 & 0.80 & 1.30 & 2.20 & 1.10 & 1.19\\
				$^\ast \text{GO-SLAM}$ \cite{14} & 0.34 & \multicolumn{1}{>{\columncolor{c2}}c}{0.29} & \multicolumn{1}{>{\columncolor{c3}}c}{0.29} & \multicolumn{1}{>{\columncolor{c2}}c}{0.32} & 0.30 & \multicolumn{1}{>{\columncolor{c3}}c}{0.39} & 0.39 & \multicolumn{1}{>{\columncolor{c3}}c}{0.46} & \multicolumn{1}{>{\columncolor{c3}}c}{0.35}\\
                \hdashline
                \multicolumn{10}{>{\columncolor{c4}}l}{\textit{3DGS-based}}\\
                SplaTAM \cite{8} & \multicolumn{1}{>{\columncolor{c3}}c}{0.31} & \multicolumn{1}{>{\columncolor{c3}}c}{0.40} & \multicolumn{1}{>{\columncolor{c3}}c}{0.29} & 0.47 & \multicolumn{1}{>{\columncolor{c3}}c}{0.27} & \multicolumn{1}{>{\columncolor{c1}}c}{\textbf{0.29}} & \multicolumn{1}{>{\columncolor{c3}}c}{0.32} & 0.72 & 0.38\\
                Gaussian-SLAM \cite{9} & \multicolumn{1}{>{\columncolor{c2}}c}{0.29} & \multicolumn{1}{>{\columncolor{c2}}c}{0.29} & \multicolumn{1}{>{\columncolor{c2}}c}{0.22} & \multicolumn{1}{>{\columncolor{c3}}c}{0.37} & \multicolumn{1}{>{\columncolor{c2}}c}{0.23} & 0.41 & \multicolumn{1}{>{\columncolor{c2}}c}{0.30} & \multicolumn{1}{>{\columncolor{c2}}c}{0.35} & \multicolumn{1}{>{\columncolor{c2}}c}{0.31}\\
                $^\ast \text{Photo-SLAM}$ \cite{17} & 0.54 & 0.39 & 0.31 & 0.52 & 0.44 & 1.28 & 0.78 & 0.58 & 0.60\\
                \textbf{GLC-SLAM (Ours)} & \multicolumn{1}{>{\columncolor{c1}}c}{\textbf{0.20}} & \multicolumn{1}{>{\columncolor{c1}}c}{\textbf{0.19}} & \multicolumn{1}{>{\columncolor{c1}}c}{\textbf{0.13}} & \multicolumn{1}{>{\columncolor{c1}}c}{\textbf{0.31}} & \multicolumn{1}{>{\columncolor{c1}}c}{\textbf{0.13}} & \multicolumn{1}{>{\columncolor{c2}}c}{0.32} & \multicolumn{1}{>{\columncolor{c1}}c}{\textbf{0.21}} & \multicolumn{1}{>{\columncolor{c1}}c}{\textbf{0.33}} & \multicolumn{1}{>{\columncolor{c1}}c}{\textbf{0.23}}\\
				\bottomrule
		\end{tabular}}
	\end{center}
\end{table}

\begin{table}[ht]
\setlength\tabcolsep{3pt}
    \begin{center}
        \caption{\textbf{Tracking Performance on TUM-RGBD \cite{2}}. LC indicates loop closure.}\label{tab:Table 2.}
        \resizebox{\linewidth}{!}{
            \begin{tabular}{lccccc}
                \toprule
                \text{Method} & \text{LC} & \texttt{fr1/desk} & \texttt{fr2/xyz} & \texttt{fr3/off.} & \text{Avg.} 
                \\
                \midrule
                \multicolumn{6}{>{\columncolor{c4}}l}{\textit{NeRF-based}}\\
                NICE-SLAM \cite{10} & \color{Red}{\usym{2717}} & 4.26 & 6.19 & 3.87 & 4.77\\
                Vox-Fusion \cite{11} & \color{Red}{\usym{2717}} & 3.52 & 1.49 & 26.01 & 10.34\\
                ESLAM \cite{7} & \color{Red}{\usym{2717}} & \multicolumn{1}{>{\columncolor{c2}}c}{2.47} & \multicolumn{1}{>{\columncolor{c2}}c}{1.11} & \multicolumn{1}{>{\columncolor{c2}}c}{2.42} & \multicolumn{1}{>{\columncolor{c2}}c}{2.00}\\
                Point-SLAM \cite{12} & \color{Red}{\usym{2717}} & 4.34 & 1.31 & 3.48 & 3.04\\
                MIPS-Fusion \cite{13} & \color{OliveGreen}{\usym{2713}} & 3.00 & 1.40 & 4.60 & 3.00\\
                \hdashline
                \multicolumn{6}{>{\columncolor{c4}}l}{\textit{3DGS-based}}\\
                SplaTAM \cite{8} & \color{Red}{\usym{2717}} & 3.35 & \multicolumn{1}{>{\columncolor{c3}}c}{1.24} & 5.16 & 3.25\\
                Gaussian-SLAM \cite{9} & \color{Red}{\usym{2717}} & 2.73 & 1.39 & 5.31 & 3.14\\
                $^\ast \text{Photo-SLAM}$ \cite{17} & \color{OliveGreen}{\usym{2713}} & \multicolumn{1}{>{\columncolor{c3}}c}{2.60} & \multicolumn{1}{>{\columncolor{c1}}c}{\textbf{0.35}} & \multicolumn{1}{>{\columncolor{c1}}c}{\textbf{1.00}} & \multicolumn{1}{>{\columncolor{c1}}c}{\textbf{1.32}}\\
                \textbf{GLC-SLAM (Ours)} & \color{OliveGreen}{\usym{2713}} & \multicolumn{1}{>{\columncolor{c1}}c}{\textbf{1.85}} & 1.30 & \multicolumn{1}{>{\columncolor{c3}}c}{3.53} & \multicolumn{1}{>{\columncolor{c3}}c}{2.23}\\
                \bottomrule
        \end{tabular}}
    \end{center}
\end{table}

\begin{table}[ht]
\setlength\tabcolsep{3pt}
	\begin{center}
		\caption{\textbf{Tracking Performance on Scannet \cite{3}}.}\label{tab:Table 3.}
		\resizebox{\linewidth}{!}{
			\begin{tabular}{lcccccccc}
				\toprule
				\texttt{Scene ID} & \texttt{0000} & \texttt{0059} & \texttt{0106} & \texttt{0169} & \texttt{0181} & \texttt{0207} & \text{Avg.}\\
				\midrule
                \multicolumn{9}{>{\columncolor{c4}}l}{\textit{NeRF-based}}\\
				NICE-SLAM \cite{10} & 12.0 & 14.0 & 7.9 & 10.9 & 13.4 & \multicolumn{1}{>{\columncolor{c2}}c}{6.2} & 10.7\\
			Vox-Fusion \cite{11} & 16.6 & 24.2 & 8.4 & 27.3 & 23.3 & 9.4 & 18.2\\
				ESLAM \cite{7} & \multicolumn{1}{>{\columncolor{c2}}c}{7.3} & 8.5 & \multicolumn{1}{>{\columncolor{c3}}c}{7.5} & \multicolumn{1}{>{\columncolor{c1}}c}{\textbf{6.5}} & \multicolumn{1}{>{\columncolor{c2}}c}{9.0} & \multicolumn{1}{>{\columncolor{c1}}c}{\textbf{5.7}} & \multicolumn{1}{>{\columncolor{c2}}c}{7.4}\\
				Point-SLAM \cite{12} & 10.2 & \multicolumn{1}{>{\columncolor{c2}}c}{7.8} & 8.7 & 22.2 & 14.8 & 9.5 & 12.2\\
                MIPS-Fusion \cite{13} & \multicolumn{1}{>{\columncolor{c3}}c}{7.9} & 10.7 & 9.7 & \multicolumn{1}{>{\columncolor{c3}}c}{9.7} & 14.2 & 7.8 & 10.0\\
				$^\ast \text{GO-SLAM}$ \cite{14} & \multicolumn{1}{>{\columncolor{c1}}c}{\textbf{5.4}} & \multicolumn{1}{>{\columncolor{c1}}c}{\textbf{7.5}} & \multicolumn{1}{>{\columncolor{c2}}c}{7.0} & \multicolumn{1}{>{\columncolor{c2}}c}{7.7} & \multicolumn{1}{>{\columncolor{c1}}c}{\textbf{6.8}} & 6.9 & \multicolumn{1}{>{\columncolor{c1}}c}{\textbf{6.9}}\\
                \hdashline
                \multicolumn{9}{>{\columncolor{c4}}l}{\textit{3DGS-based}}\\
                MonoGS \cite{16} & 9.8 & 32.1 & 8.9 & 10.7 & 21.8 & 7.9 & 15.2\\
                SplaTAM \cite{8} & 12.8 & 10.1 & 17.7 & 12.1 & 11.1 & 7.5 & 11.9\\
                Gaussian-SLAM \cite{9} & 24.8 & 12.8 & 13.5 & 16.3 & 21.0 & 14.3 & 17.1\\
                \textbf{GLC-SLAM (Ours)} & 12.9 & \multicolumn{1}{>{\columncolor{c3}}c}{7.9} & \multicolumn{1}{>{\columncolor{c1}}c}{\textbf{6.3}} & 10.5 & \multicolumn{1}{>{\columncolor{c3}}c}{11.0} & \multicolumn{1}{>{\columncolor{c3}}c}{6.3} & \multicolumn{1}{>{\columncolor{c3}}c}{9.2}\\
				\bottomrule
		\end{tabular}}
	\end{center}
\end{table}

\begin{table}[ht]
\setlength\tabcolsep{3pt}
	\begin{center}
		\caption{\textbf{Rendering Performance on Replica \cite{1}}.}\label{tab:Table 4.}
		\resizebox{\linewidth}{!}{
			\begin{tabular}{lccccc}
				\toprule
			  \multirow{2}*{Metric} & ESLAM\cite{7} & Point- & SplaTAM\cite{8} & Photo- & \textbf{Ours}\\
                ~ & & SLAM\cite{12} & & SLAM \cite{1} & \\
				\midrule
                PSNR $\uparrow$ & 27.80 & \multicolumn{1}{>{\columncolor{c2}}c}{35.17} & 34.11 & \multicolumn{1}{>{\columncolor{c3}}c}{34.96} & \multicolumn{1}{>{\columncolor{c1}}c}{\textbf{41.07}}\\
                SSIM $\uparrow$ & 0.921 & \multicolumn{1}{>{\columncolor{c2}}c}{0.975} & \multicolumn{1}{>{\columncolor{c3}}c}{0.970} & 0.942 & \multicolumn{1}{>{\columncolor{c1}}c}{\textbf{0.995}}\\
                LPIPS $\downarrow$ & 0.245 & 0.124  & \multicolumn{1}{>{\columncolor{c3}}c}{0.100} & \multicolumn{1}{>{\columncolor{c2}}c}{0.059} & \multicolumn{1}{>{\columncolor{c1}}c}{\textbf{0.021}}\\
				\bottomrule
		\end{tabular}}
	\end{center}
\end{table}

\section{EXPERIMENT}
\subsection{Experimental Setup} We describe our experimental setup and evaluate our method against state-of-the-art dense RGB-D SLAM methods on Replica \cite{1} as well as the real world TUM-RGBD \cite{2} and the ScanNet \cite{3} datasets.

\subsubsection{Datasets} The Replica dataset \cite{1} consists of high-quality 3D reconstructions of diverse indoor scenes. We leverage the publicly available dataset by Sucar \textit{et al.} \cite{4}, which contains trajectories from an RGB-D sensor. Additionally, we showcase our framework on real-world data using the TUM-RGBD dataset \cite{2} and the ScanNet dataset \cite{3}. The TUM-RGBD poses were captured utilizing an external motion capture system, while ScanNet uses poses from BundleFusion \cite{5}.

\subsubsection{Metrics} We evaluate camera tracking accuracy using ATE RMSE \cite{2}. Rendering quality is evaluated by comparing full-resolution rendered images to input training views using peak signal-to-noise ratio (PSNR), SSIM \cite{24}, and LPIPS \cite{26} metrics. Reconstruction performance is measured on meshes produced by marching cubes \cite{6} using the F1-score, which is the harmonic mean of the Precision (P) and Recall (R). We also report the depth L1 metric, which compares mesh depth at random poses to its ground truth.

\subsubsection{Baseline Methods} We primarily compare our method to existing state-of-the-art dense RGB-D SLAM methods such as ESLAM \cite{7}, GO-SLAM \cite{14}, SplaTAM \cite{8} and Gaussian-SLAM \cite{9}. We use the reported numbers from the respective papers where available and for others, we reproduce the results by running the official code.

\subsubsection{Implementation details} We run GLC-SLAM on a desktop PC with an Intel Core i9-12900KF CPU and an NVIDIA RTX 3090 GPU. In all our experiments, we set alpha threshold $\alpha_\text{thre}=0.6$ and the local loop detection threshold $s_\text{local}=0.8$. For submap optimization, we select $k=5$ keyframes with $\lambda_\text{color},\text{ }\lambda_\text{depth} \text{ and } \lambda_\text{reg}$ to 1.

\begin{figure*}[ht]
    \centering
\begin{minipage}[c]{0.03\linewidth}
    \centering
    \rotatebox{90}{}
\end{minipage}
\begin{minipage}[c]{0.18\linewidth}
    \centering\small\text{Point-SLAM \cite{12}}
\end{minipage}
\begin{minipage}[c]{0.18\linewidth}
    \centering\small\text{GO-SLAM \cite{14}}
\end{minipage}
\begin{minipage}[c]{0.18\linewidth}
    \centering\small\text{Gaussian-SLAM \cite{9}}
\end{minipage}
\begin{minipage}[c]{0.18\linewidth}
    \centering\small\textbf{GLC-SLAM (Ours)}
\end{minipage}
\begin{minipage}[c]{0.18\linewidth}
    \centering\small\text{Ground Truth}
\end{minipage}

\centering
\begin{minipage}[c]{0.03\linewidth}
    \centering
    \rotatebox{90}{\small\texttt{scene 0207}}
\end{minipage}
\begin{minipage}[c]{0.18\linewidth}
    {\epsfig{figure=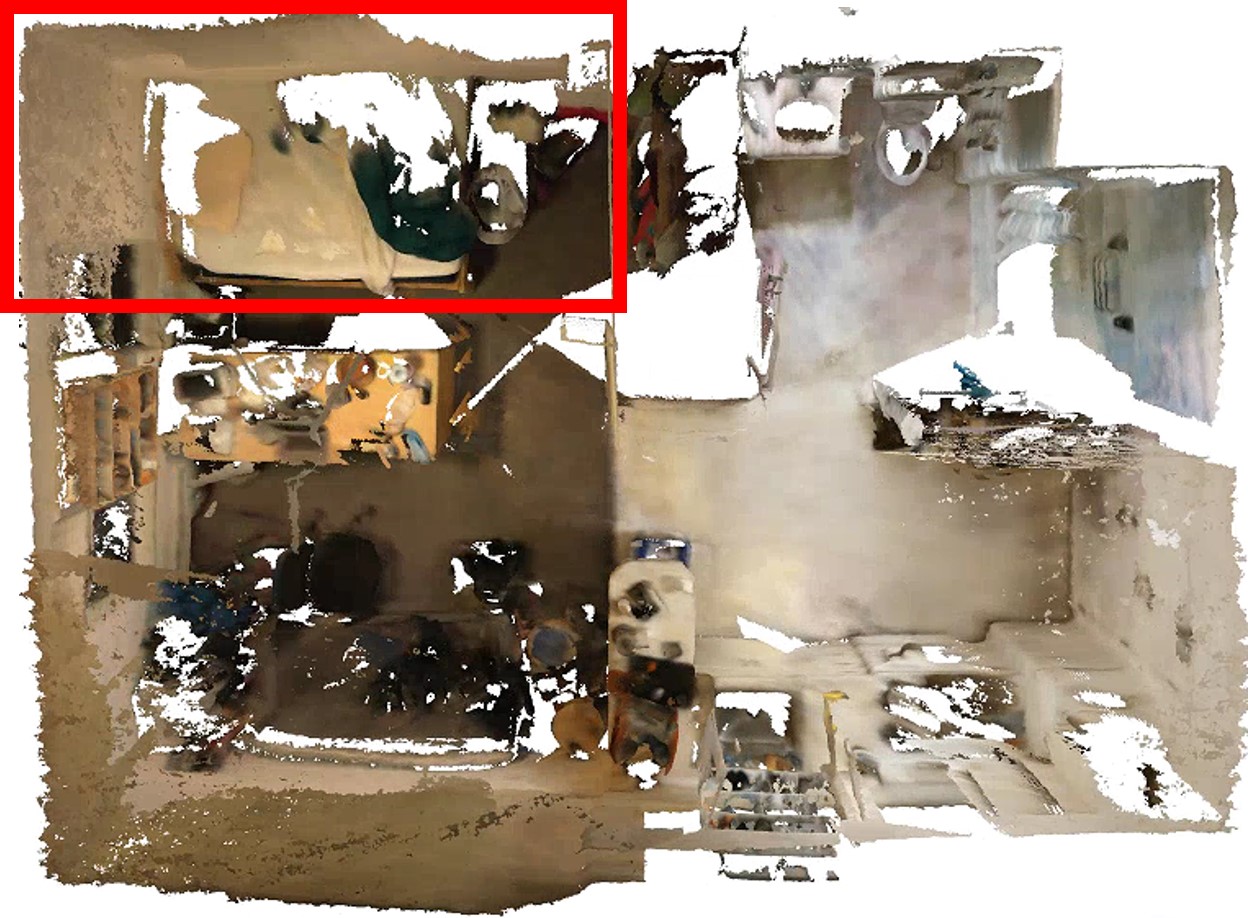,width=3cm}}
\end{minipage}
\begin{minipage}[c]{0.18\linewidth}
    {\epsfig{figure=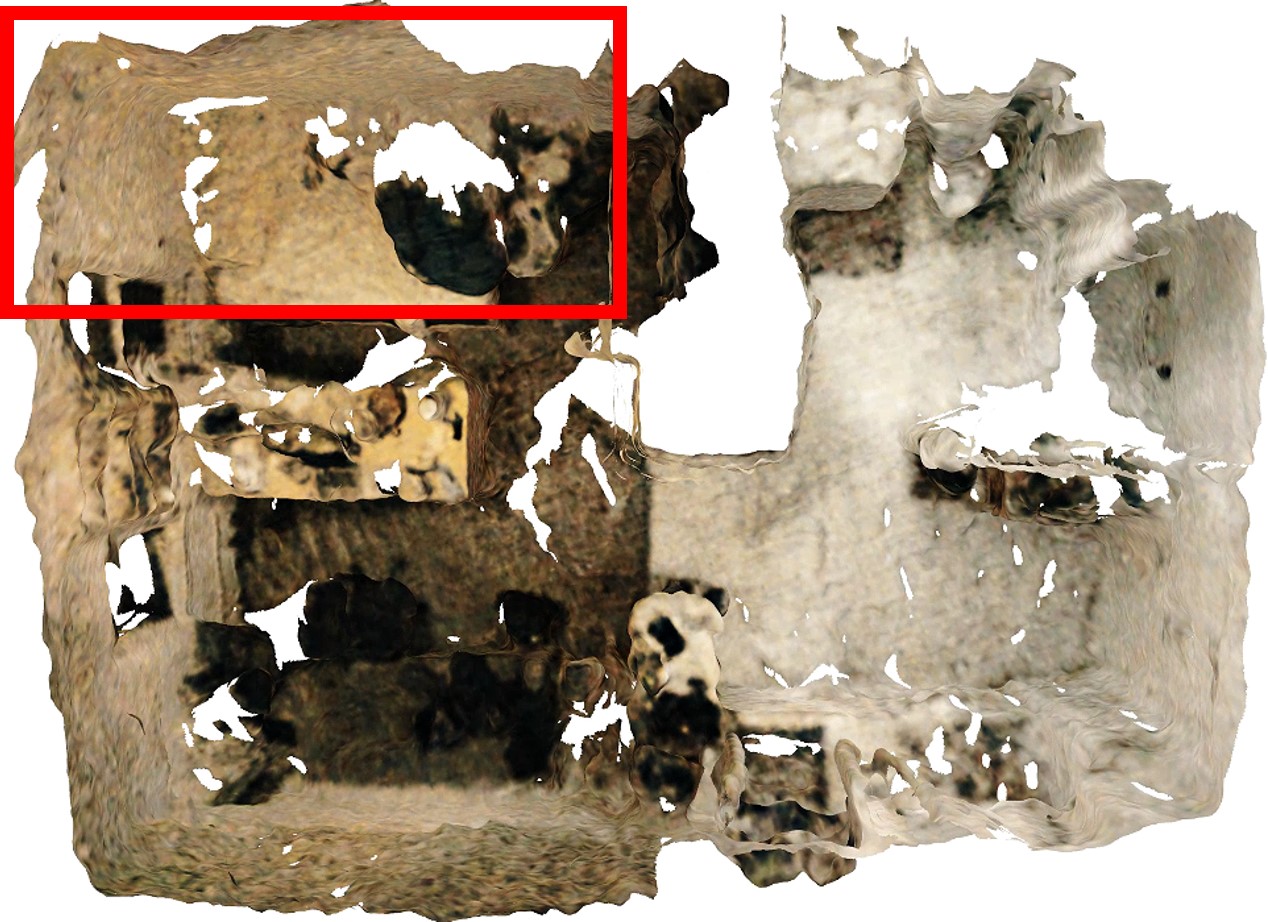,width=3cm}}
\end{minipage}
\begin{minipage}[c]{0.18\linewidth}
    {\epsfig{figure=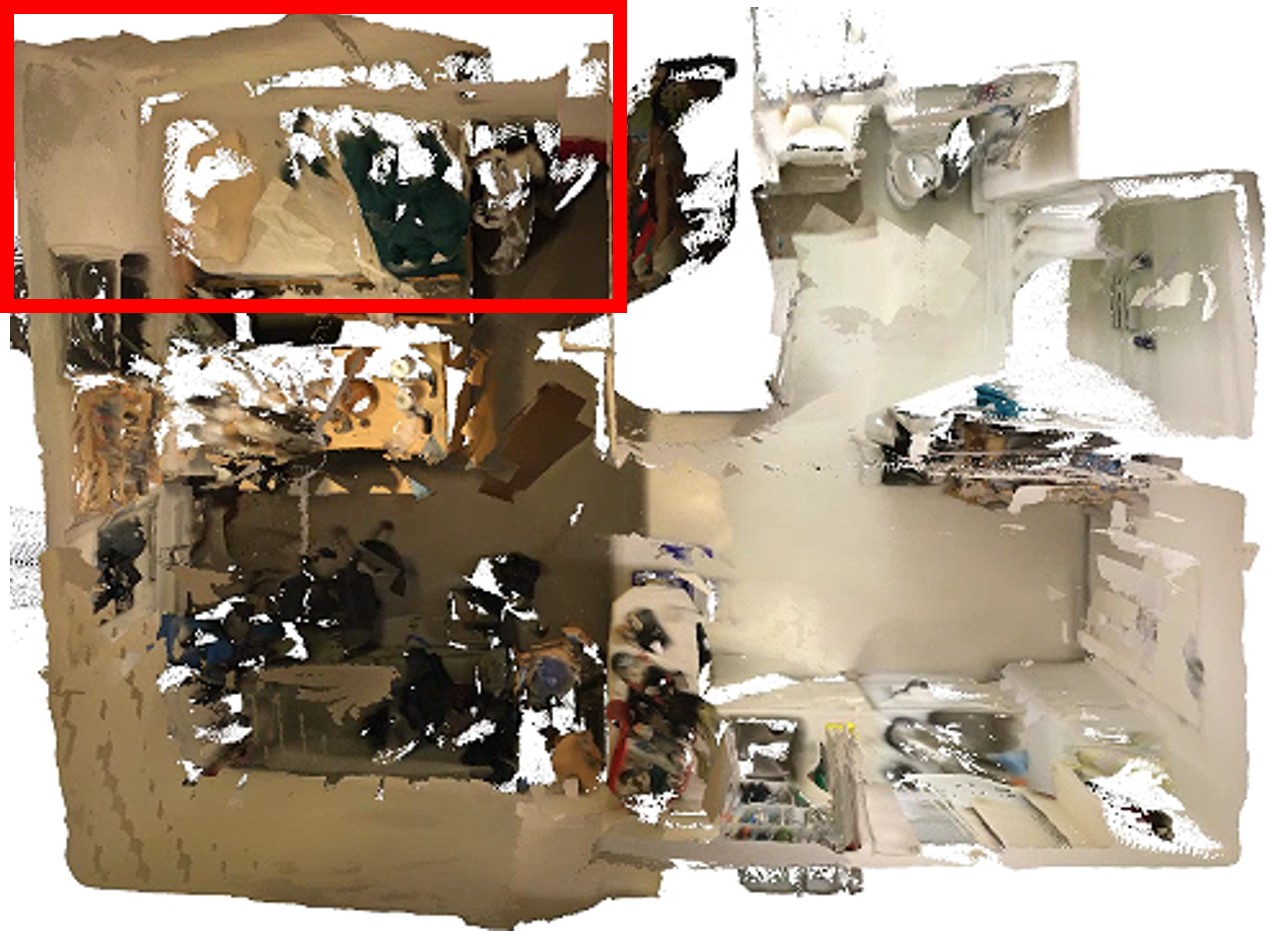,width=3cm}}
\end{minipage}
\begin{minipage}[c]{0.18\linewidth}
    {\epsfig{figure=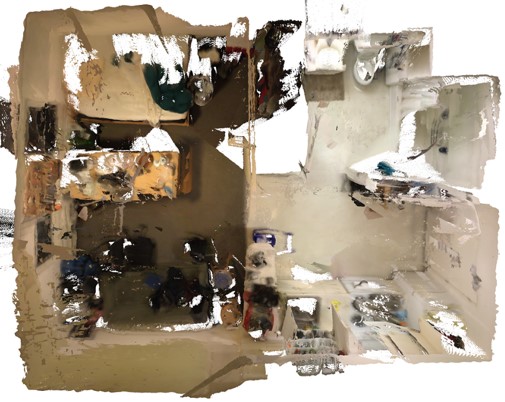,width=3cm}}
\end{minipage}
\begin{minipage}[c]{0.18\linewidth}
    {\epsfig{figure=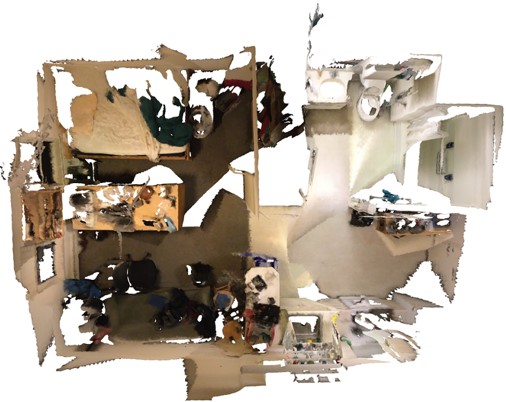,width=3cm}}
\end{minipage}

\centering
\begin{minipage}[c]{0.03\linewidth}
    \centering
    \rotatebox{90}{\small\texttt{scene 0181}}
\end{minipage}%
\begin{minipage}[c]{0.18\linewidth}
    \centering
    {\epsfig{figure=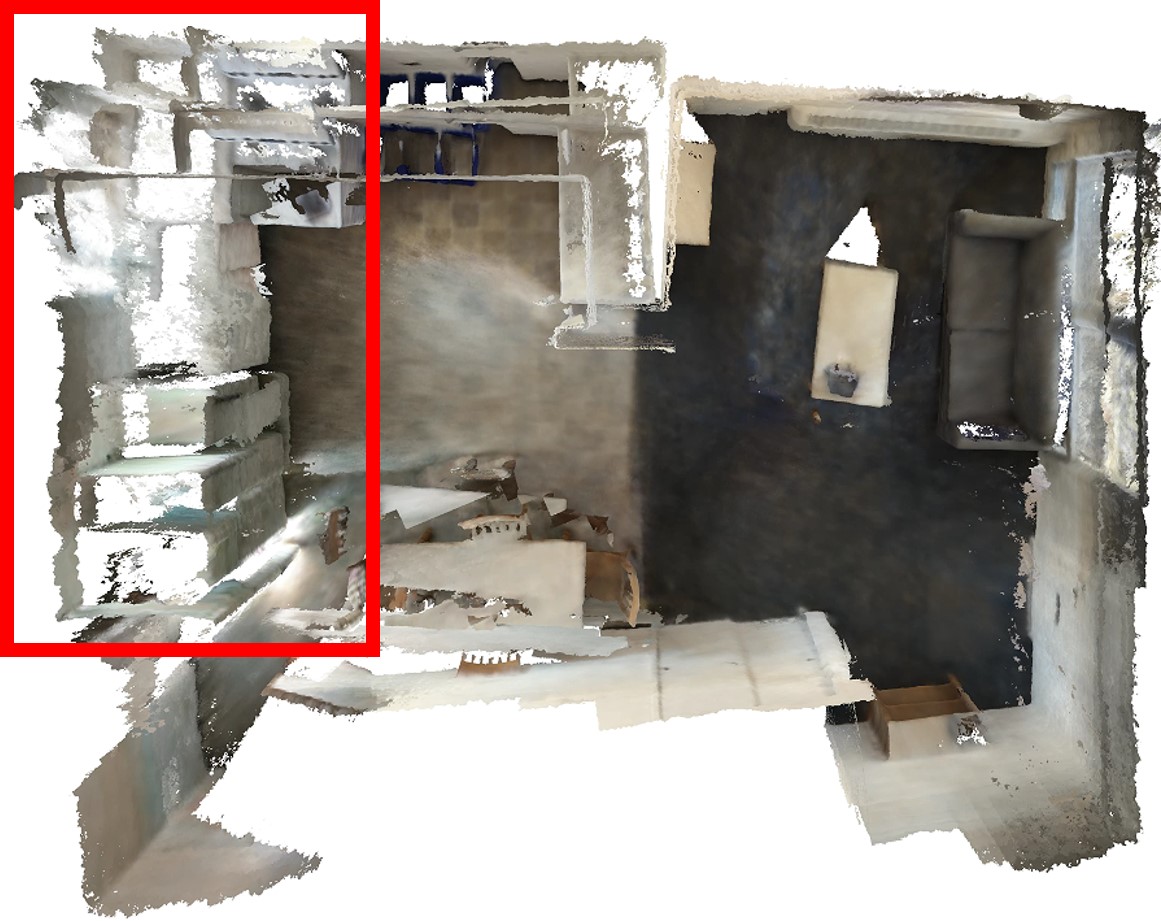,width=3cm}}
\end{minipage}
\begin{minipage}[c]{0.18\linewidth}
    \centering
    {\epsfig{figure=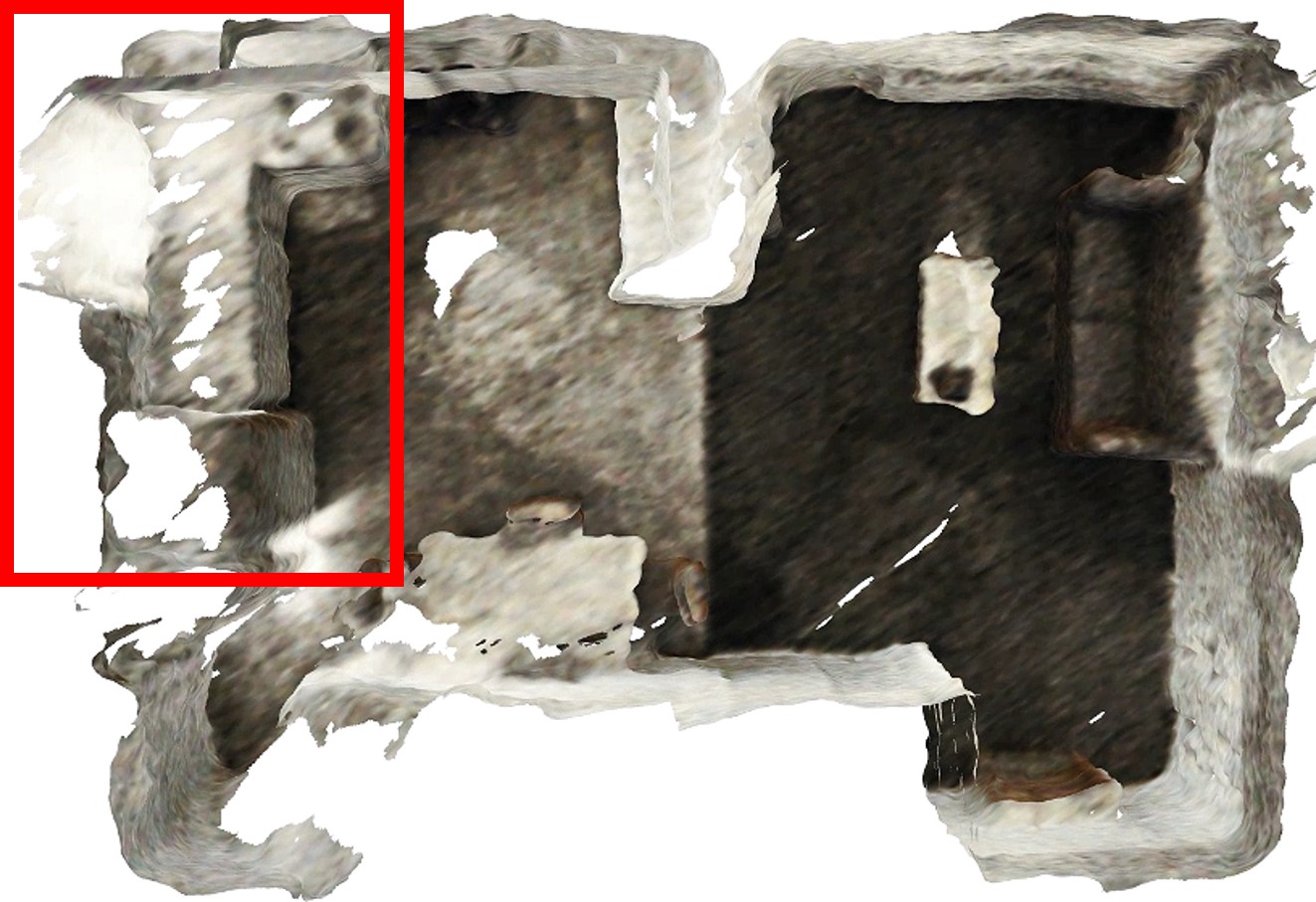,width=3cm}}
\end{minipage}
\begin{minipage}[c]{0.18\linewidth}
    \centering
    {\epsfig{figure=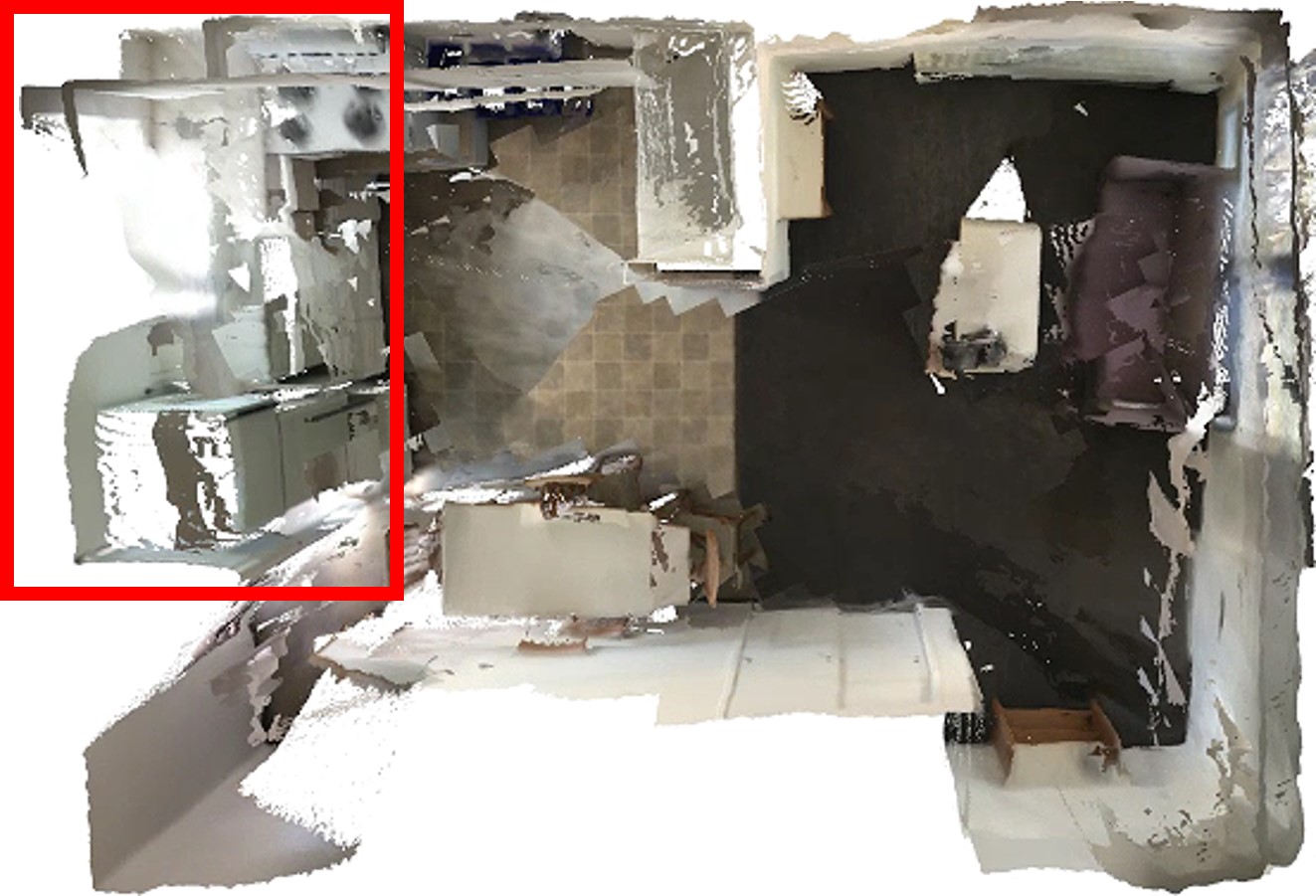,width=3cm}}
\end{minipage}
\begin{minipage}[c]{0.18\linewidth}
    \centering
    {\epsfig{figure=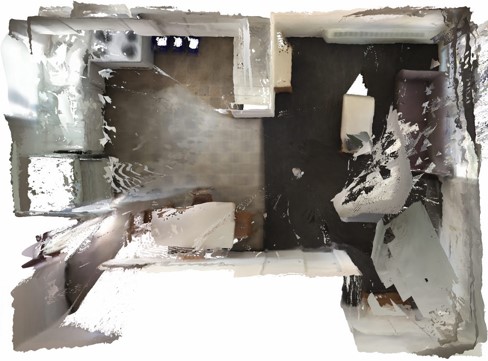,width=3cm}}
\end{minipage}
\begin{minipage}[c]{0.18\linewidth}
    \centering
    {\epsfig{figure=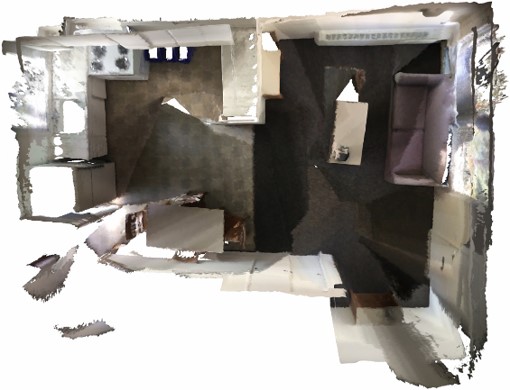,width=3cm}}
\end{minipage}
\centering
\begin{minipage}[c]{0.03\linewidth}

    \centering
    \rotatebox{90}{\small\texttt{scene 0169}}
\end{minipage}%
\begin{minipage}[c]{0.18\linewidth}
    \centering
    {\epsfig{figure=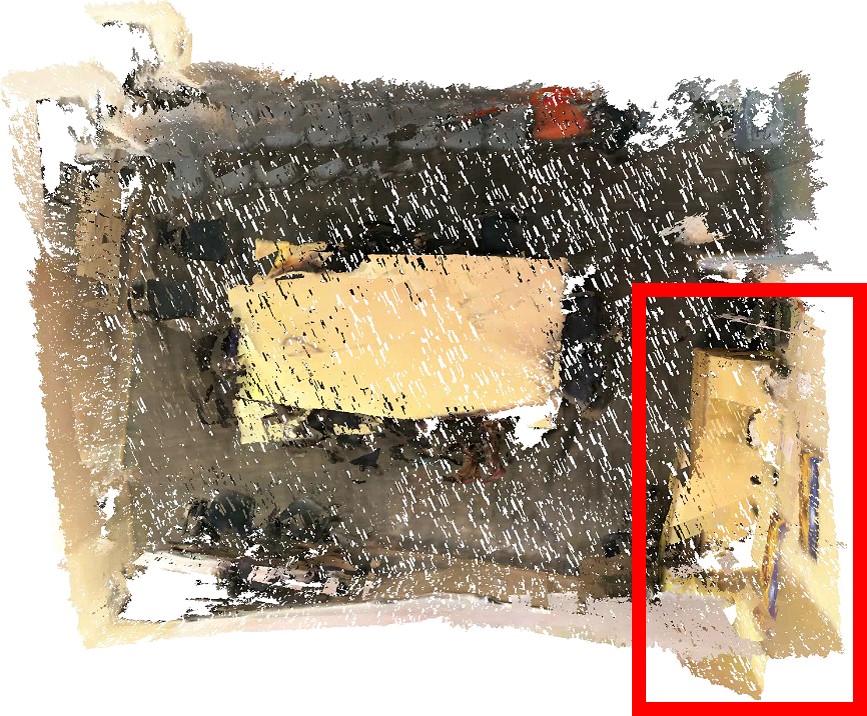,width=3cm}}
\end{minipage}
\begin{minipage}[c]{0.18\linewidth}
    \centering
    {\epsfig{figure=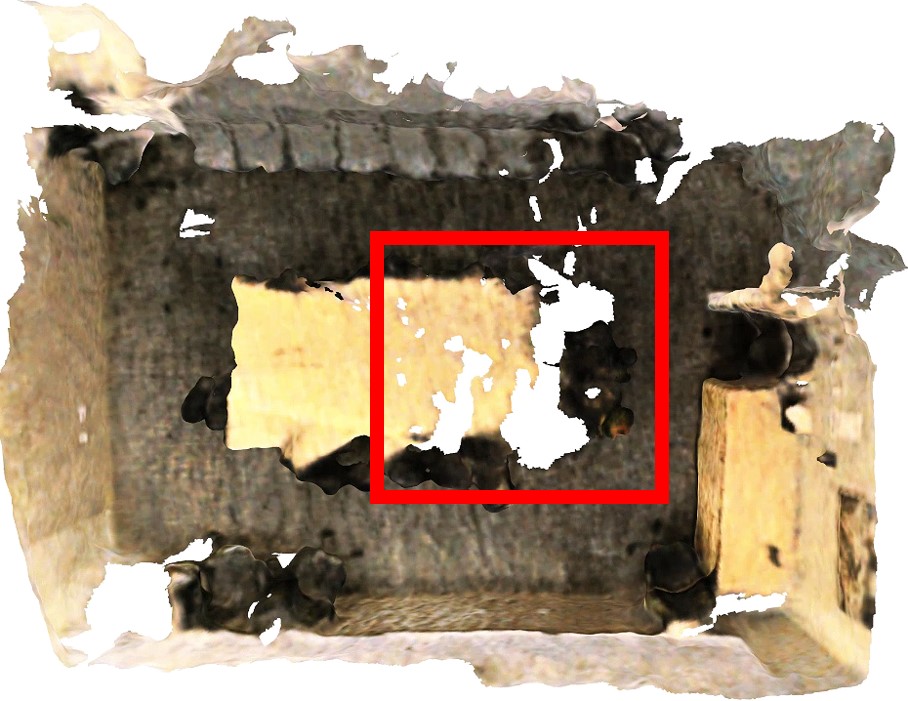,width=3cm}}
\end{minipage}
\begin{minipage}[c]{0.18\linewidth}
    \centering
    {\epsfig{figure=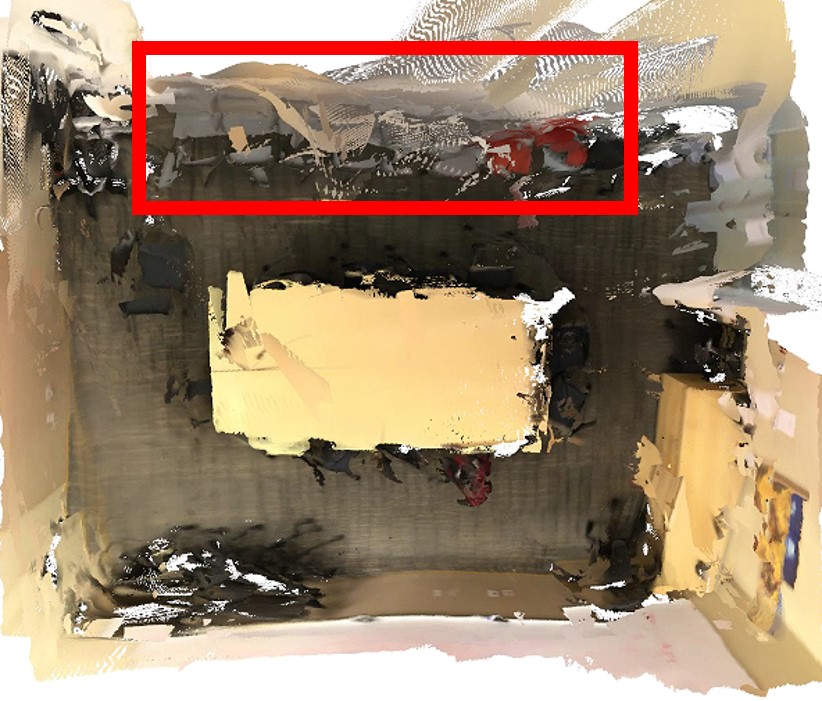,width=3cm}}
\end{minipage}
\begin{minipage}[c]{0.18\linewidth}
    \centering
    {\epsfig{figure=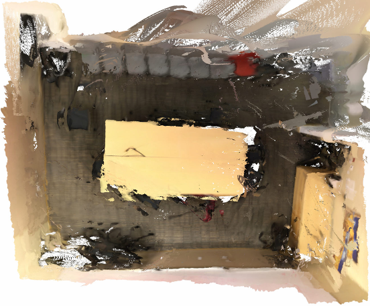,width=3cm}}
\end{minipage}
\begin{minipage}[c]{0.18\linewidth}
    \centering
    {\epsfig{figure=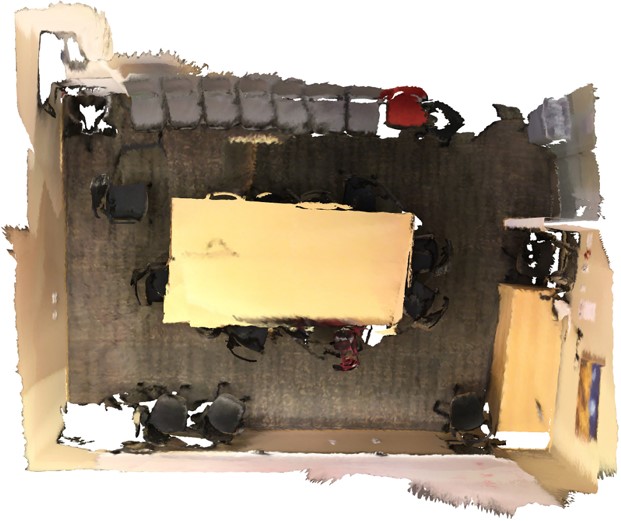,width=3cm}}
\end{minipage}
\caption{\textbf{Mesh Evaluation on ScanNet \cite{3}}. The red boxes show map drift or poor details.}
\label{fig:Figure3}
\end{figure*} 
\subsection{Tracking Evaluation}
We report the camera tracking performance in Tables \ref{tab:Table 1.} to \ref{tab:Table 3.}. On the Replica dataset, our approach outperforms all competing techniques, 
achieving a 26\% improvement in average accuracy over the second-best method.
On the TUM-RGBD dataset, our method surpasses all 3DGS-based approaches except Photo-SLAM \cite{17}, which incorporates ORB-SLAM \cite{18} tracker.
On the ScanNet dataset, our method achieves the highest pose accuracy among all 3DGS-based baselines, showing the effectiveness of our proposed loop closure strategy in reducing accumulated tracking errors in real-world environments.

\subsection{Mapping Evaluation}
Tab. \ref{tab:Table 4.} compares rendering performance on the Replica dataset and shows that our approach achieves superior results on all three evaluated metrics compared to competing methods. We evaluate the reconstruction performance on both ScanNet and Replica. As shown in Fig. \ref{fig:Figure3}, our method accurately recovers geometric details and mitigates map drift as highlighted in red boxes, especially in edge areas. In Tab. \ref{tab:Table 5.}, we present a quantitative comparison where GLC-SLAM shows competitive performance against 3GDS-based methods but falls behind NeRF-based methods due to its limited hole-filling capability.

\subsection{Runtime and Memory Analysis}
In Tab. \ref{tab:Table 6.} we compare runtime and memory usage on the Replica \texttt{office0} scene. We report both per-iteration and per-frame runtime profiled on a RTX 3090 GPU. 
Our method achieves the fastest per-iteration and comparable per-frame running speed while maintaining the lowest GPU memory consumption. 

\begin{table}[t]
\setlength\tabcolsep{3pt}
\renewcommand{\arraystretch}{1.2}
	\begin{center}
		\caption{\textbf{Reconstruction Performance on Replica \cite{1}}.}\label{tab:Table 5.}
		\resizebox{\linewidth}{!}{
			\begin{tabular}{llccccccccc}
				\toprule
				\text{Method} & \text{Metric} & \texttt{rm0} & \texttt{rm1} & \texttt{rm2} & \texttt{off0} 
                & \texttt{off1} & \texttt{off2} & \texttt{off3} & \texttt{off4} & \text{Avg.} 
				\\
				\midrule
                \multicolumn{11}{>{\columncolor{c4}}l}{\textit{NeRF-based}}\\
				NICE- & Depth L1 $[$cm$]\downarrow$ & 1.81 & 1.44 & 2.04 & 1.39 & 1.76 & 8.33 & 4.99 & 2.01 & 2.97\\
                SLAM \cite{10} & \text{F1 $[\%]\uparrow$} & 45.0 & 44.8 & 43.6 & 50.0 & 51.9 & 39.2 & 39.9 & 36.5 & 43.9\\
                \hdashline
                Vox- & Depth L1 $[$cm$]\downarrow$ & 1.09 & 1.90 & 2.21 & 2.32 & 3.40 & 4.19 & 2.96 & 1.61 & 2.46\\
                Fusion \cite{11} & \text{F1 $[\%]\uparrow$} & 17.3 & 33.4 & 24.0 & 43.0 & 31.8 & 21.8 & 17.3 & 22.0 & 26.3\\
                \hdashline
                \multirow{2}*{ESLAM \cite{7}} & Depth L1 $[$cm$]\downarrow$ & 0.97 & 1.07 & 1.28 & 0.86 & 1.26 & 1.71 & 1.43 & 1.06 & 1.18\\
                ~ & \text{F1 $[\%]\uparrow$} & 81.0 & 82.2 & 83.9 & 78.4 & 75.5 & 77.1 & 75.5 & 79.1 & 79.1\\
                \hdashline
                Point- & Depth L1 $[$cm$]\downarrow$ & \multicolumn{1}{>{\columncolor{c2}}c}{0.53} & \multicolumn{1}{>{\columncolor{c1}}c}{\textbf{0.22}} & \multicolumn{1}{>{\columncolor{c1}}c}{\textbf{0.46}} & \multicolumn{1}{>{\columncolor{c1}}c}{\textbf{0.30}} & \multicolumn{1}{>{\columncolor{c3}}c}{0.57} & \multicolumn{1}{>{\columncolor{c1}}c}{\textbf{0.49}} & \multicolumn{1}{>{\columncolor{c1}}c}{\textbf{0.51}} & \multicolumn{1}{>{\columncolor{c3}}c}{0.46} & \multicolumn{1}{>{\columncolor{c1}}c}{\textbf{0.44}}\\
                SLAM \cite{12} & \text{F1 $[\%]\uparrow$} & \multicolumn{1}{>{\columncolor{c3}}c}{86.9} & \multicolumn{1}{>{\columncolor{c1}}c}{\textbf{92.3}} & \multicolumn{1}{>{\columncolor{c1}}c}{\textbf{90.8}} & \multicolumn{1}{>{\columncolor{c1}}c}{\textbf{93.8}} & \multicolumn{1}{>{\columncolor{c1}}c}{\textbf{91.6}} & \multicolumn{1}{>{\columncolor{c1}}c}{\textbf{89.0}} & \multicolumn{1}{>{\columncolor{c1}}c}{\textbf{88.2}} & \multicolumn{1}{>{\columncolor{c3}}c}{85.6} & \multicolumn{1}{>{\columncolor{c1}}c}{\textbf{89.8}}\\
                \hdashline
                GO- & $ \text{Depth L1} [\text{cm}]\downarrow$ & 4.56 & 1.97 & 3.43 & 2.47 & 3.03 & 10.3 & 7.31 & 4.34 & 4.68\\
                SLAM \cite{14} & \text{F1 $[\%]\uparrow$} & 69.9 & 34.4 & 59.7 & 46.5 & 40.8 & 51.0 & 64.6 & 50.7 & 52.2\\
                \hdashline
                \multicolumn{11}{>{\columncolor{c4}}l}{\textit{3DGS-based}}\\
                \multirow{2}*{SplaTAM \cite{8}} & Depth L1 $[$cm$]\downarrow$ & \multicolumn{1}{>{\columncolor{c1}}c}{\textbf{0.43}} & 0.38 & \multicolumn{1}{>{\columncolor{c3}}c}{0.54} & \multicolumn{1}{>{\columncolor{c2}}c}{0.44} & 0.66 & \multicolumn{1}{>{\columncolor{c3}}c}{1.05} & \multicolumn{1}{>{\columncolor{c2}}c}{1.60} & 0.68 & 0.72\\
                ~ & \text{F1 $[\%]\uparrow$} & \multicolumn{1}{>{\columncolor{c1}}c}{\textbf{89.3}} & 88.2 & \multicolumn{1}{>{\columncolor{c3}}c}{88.0} & \multicolumn{1}{>{\columncolor{c3}}c}{91.7} & \multicolumn{1}{>{\columncolor{c3}}c}{90.0} & 85.1 & 77.1 & 80.1 & 86.1\\
                \hdashline
                Gaussian- & Depth L1 $[$cm$]\downarrow$ & 0.61  & \multicolumn{1}{>{\columncolor{c3}}c}{0.25} & \multicolumn{1}{>{\columncolor{c3}}c}{0.54} & \multicolumn{1}{>{\columncolor{c3}}c}{0.50} & \multicolumn{1}{>{\columncolor{c2}}c}{0.52} & \multicolumn{1}{>{\columncolor{c2}}c}{0.98} & \multicolumn{1}{>{\columncolor{c3}}c}{1.63} & \multicolumn{1}{>{\columncolor{c1}}c}{\textbf{0.42}} & \multicolumn{1}{>{\columncolor{c2}}c}{0.68}\\
                SLAM \cite{9} & \text{F1 $[\%]\uparrow$} & \multicolumn{1}{>{\columncolor{c2}}c}{88.8} & \multicolumn{1}{>{\columncolor{c2}}c}{91.4} & \multicolumn{1}{>{\columncolor{c2}}c}{90.5} & \multicolumn{1}{>{\columncolor{c3}}c}{91.7} & \multicolumn{1}{>{\columncolor{c2}}c}{90.1} & \multicolumn{1}{>{\columncolor{c3}}c}{87.3} & \multicolumn{1}{>{\columncolor{c3}}c}{84.2} & \multicolumn{1}{>{\columncolor{c1}}c}{\textbf{87.4}} & \multicolumn{1}{>{\columncolor{c3}}c}{88.9}\\
                \hdashline
                \multirow{2}*{\textbf{Ours}} & Depth L1 $[$cm$]\downarrow$ & \multicolumn{1}{>{\columncolor{c3}}c}{0.57} & \multicolumn{1}{>{\columncolor{c2}}c}{0.24} & \multicolumn{1}{>{\columncolor{c2}}c}{0.50} & \multicolumn{1}{>{\columncolor{c2}}c}{0.44} & \multicolumn{1}{>{\columncolor{c1}}c}{\textbf{0.48}} & 1.06 & 1.85 & \multicolumn{1}{>{\columncolor{c2}}c}{0.45} & \multicolumn{1}{>{\columncolor{c3}}c}{0.70}\\
                ~ & \text{F1 $[\%]\uparrow$} & \multicolumn{1}{>{\columncolor{c1}}c}{\textbf{89.3}} & \multicolumn{1}{>{\columncolor{c3}}c}{91.3} & \multicolumn{1}{>{\columncolor{c2}}c}{90.5} & \multicolumn{1}{>{\columncolor{c2}}c}{92.3} & \multicolumn{1}{>{\columncolor{c3}}c}{90.0} & \multicolumn{1}{>{\columncolor{c2}}c}{87.7} & \multicolumn{1}{>{\columncolor{c2}}c}{84.4} & \multicolumn{1}{>{\columncolor{c2}}c}{87.3} & \multicolumn{1}{>{\columncolor{c2}}c}{89.1}\\
				\bottomrule
		\end{tabular}}
	\end{center}
\end{table}

\subsection{Ablation Study}
In Tab. \ref{tab:Table 7.} we ablate the effectiveness of loop closure and keyframe selection for the tracking and mapping performance on Replica \texttt{room0} scene. The results indicate that the absence of loop closure significantly degrades tracking accuracy and reduces robustness. We also test our method using random keyframe selection 
. In contrast, our uncertainty-minimized strategy enhances the optimization process by incorporating more valuable keyframes, which is crucial for achieving accurate mapping.

\begin{table}[ht]
\setlength\tabcolsep{2pt}
	\begin{center}
		\caption{\textbf{Runtime memory performance on Replica \cite{1}} \texttt{office0}.}\label{tab:Table 6.}
		\resizebox{\linewidth}{!}{
			\begin{tabular}{lcccccc}
				\toprule
				\multirow{2}*{Method} & Mapping & Mapping & Tracking & Tracking & Peak GPU\\
                ~ & /Iter(ms) & /Frame(s) & /Iter(ms) & /Frame(s) & Use(GiB)\\
				\midrule
				NICE-SLAM \cite{10} & 70 & 4.43 & \multicolumn{1}{>{\columncolor{c3}}c}{20} & 1.76 & \multicolumn{1}{>{\columncolor{c3}}c}{10.4}\\
				ESLAM \cite{7} & \multicolumn{1}{>{\columncolor{c2}}c}{36} & \multicolumn{1}{>{\columncolor{c1}}c}{\textbf{0.62}} & \multicolumn{1}{>{\columncolor{c2}}c}{17} & \multicolumn{1}{>{\columncolor{c1}}c}{\textbf{0.14}} & 17.5\\
				Point-SLAM \cite{12} & \multicolumn{1}{>{\columncolor{c3}}c}{41} & \multicolumn{1}{>{\columncolor{c3}}c}{2.56} & \multicolumn{1}{>{\columncolor{c3}}c}{20} & \multicolumn{1}{>{\columncolor{c2}}c}{0.85} & \multicolumn{1}{>{\columncolor{c2}}c}{7.3}\\
                SplaTAM \cite{8} & 80 & 4.81 & 66 & 2.65 & 10.5\\
                \textbf{GLC-SLAM (Ours)} & \multicolumn{1}{>{\columncolor{c1}}c}{\textbf{18}} & \multicolumn{1}{>{\columncolor{c2}}c}{0.80} & \multicolumn{1}{>{\columncolor{c1}}c}{\textbf{16}} & \multicolumn{1}{>{\columncolor{c3}}c}{1.07} & \multicolumn{1}{>{\columncolor{c1}}c}{\textbf{7.0}}\\
				\bottomrule
		\end{tabular}}
	\end{center}
\end{table}

\begin{table}[!ht]
	\begin{center}
		\caption{\textbf{Ablation study on Replica \cite{1}} \texttt{room0}. LC and KF indicate Loop Closure and Keyframe.}\label{tab:Table 7.}
		\resizebox{\linewidth}{!}{
            \begin{tabular}{ccccc}
            \toprule
            LC & KF Selection & ATE $[$cm$]$ & Depth L1 $[$cm$]$ & F1 $[\%]$\\
            \midrule
            \color{Red}{\usym{2717}} & \color{Red}{\usym{2717}} & 0.29 & \multicolumn{1}{>{\columncolor{c3}}c}{0.61} & 88.8\\
            \color{Red}{\usym{2717}} & \color{OliveGreen}{\usym{2713}} & \multicolumn{1}{>{\columncolor{c2}}c}{0.26} & \multicolumn{1}{>{\columncolor{c3}}c}{0.61} & \multicolumn{1}{>{\columncolor{c3}}c}{89.0}\\
            \color{OliveGreen}{\usym{2713}} & \color{Red}{\usym{2717}} & \multicolumn{1}{>{\columncolor{c3}}c}{0.27} & \multicolumn{1}{>{\columncolor{c2}}c}{0.60} & \multicolumn{1}{>{\columncolor{c2}}c}{89.1}\\
            \color{OliveGreen}{\usym{2713}} & \color{OliveGreen}{\usym{2713}} & \multicolumn{1}{>{\columncolor{c1}}c}{\textbf{0.20}} & \multicolumn{1}{>{\columncolor{c1}}c}{\textbf{0.57}} & \multicolumn{1}{>{\columncolor{c1}}c}{\textbf{89.3}}\\
            \bottomrule
        \end{tabular}}
	\end{center}
\end{table}

\section{CONCLUSIONS}
We present GLC-SLAM, a dense RGB-D SLAM system which utilizes submaps of 3D Gaussians for local mapping and tracking and a pose graph for global pose and map optimization. The proposed loop closure module efficiently reduces accumulated errors and map drift thanks to the hierarchical loop detection and rapid map updates. To further improve the robustness of submap optimization, we design a uncertainty-minimized keyframe selection strategy to select keyframes observing more informative 3D Gaussians. Our experiments show that GLC-SLAM leverages the benefit of the 3D Gaussian representation and equips it with loop closure to demonstrate superior tracking and rendering performance as well as competitive mapping accuracy on various datasets. 




\section*{ACKNOWLEDGMENT}
This work was supported by the Key R{\&}D Program of Zhejiang Province (No. 2023C01181)

\bibliographystyle{IEEEtran}
\bibliography{root}

\end{document}